\documentclass[12pt]{article}
\begin{document}

\title{\normalsize\sc Technical Report \hfill IDSIA-07-07
\vskip 2mm\bf\Large\hrule height5pt \vskip 4mm
A Collection of Definitions of Intelligence
\vskip 4mm \hrule height2pt}

\author{
{\bf Shane Legg}\\
{\normalsize IDSIA, Galleria 2, Manno-Lugano CH-6928, Switzerland}\\
{\normalsize \tt{shane@idsia.ch\ \ \ www.idsia.ch/$\sim$shane}}\\
\ \\
{\bf Marcus Hutter}\\[3mm]
{\normalsize IDSIA, Galleria 2, Manno-Lugano CH-6928, Switzerland}\\
{\normalsize RSISE/ANU/NICTA, Canberra, ACT, 0200, Australia}\\
{\normalsize \tt{marcus@hutter1.net\ \ \ www.hutter1.net}}
}
\date{15 June 2007}

\maketitle

\begin{abstract} 
This paper is a survey of a large number of informal definitions of
``intelligence'' that the authors have collected over the years.
Naturally, compiling a complete list would be impossible as many
definitions of intelligence are buried deep inside articles and
books. Nevertheless, the 70-odd definitions presented here
are, to the authors' knowledge, the largest and most well referenced
collection there is.
\def\contentsname{\centering\normalsize Contents}
{\parskip=-2.7ex\tableofcontents}
\end{abstract}

\vspace{2ex}\centerline{\bf\small Keywords}
\begin{quote}\small
Intelligence definitions,
collective,
psychologist,
artificial,
universal.
\par\end{quote}\vskip 1ex

\newpage
\section{Introduction}

\begin{quote}\it
``Viewed narrowly, there seem to be almost as many definitions of
intelligence as there were experts asked to define it.'' \par
\hfill --- {\sl R. J. Sternberg quoted in~\cite{Gregory:98}}
\end{quote}

Despite a long history of research and debate, there is still no
standard definition of intelligence.  This has lead some to believe
that intelligence may be approximately described, but cannot be fully
defined.  We believe that this degree of pessimism is too strong.
Although there is no single standard definition, if one surveys the
many definitions that have been proposed, strong similarities between
many of the definitions quickly become obvious.  In many cases
different definitions, suitably interpreted, actually say the same
thing but in different words.  This observation lead us to believe
that a single general and encompassing definition for arbitrary
systems was possible.  Indeed we have constructed a formal definition
of intelligence, called \emph{universal
  intelligence}~\cite{Legg:06ior}, which has strong connections to the
theory of optimal learning agents~\cite{Hutter:04uaibook}.

Rather than exploring very general formal definitions of intelligence,
here we will instead take the opportunity to present the many informal
definitions that we have collected over the years.  Naturally,
compiling a complete list would be impossible as many definitions of
intelligence are buried deep inside articles and books.  Nevertheless,
the 70 odd definitions presented below are, to the best of our
knowledge, the largest and most well referenced collection there~is.
We continue to add to this collect as we discover further definitions,
and keep the most up to date version of the collection available
online~\cite{Legg:idefs}.  If you know of additional definitions that
we could add, please send us an email.

\section{Collective definitions}

In this section we present definitions that have been proposed by
groups or organisations.  In many cases definitions of intelligence
given in encyclopedias have been either contributed by an individual
psychologist or quote an earlier definition given by a psychologist.
In these cases we have chosen to attribute the quote to the
psychologist, and have placed it in the next section.  In this section
we only list those definitions that either cannot be attributed to a
specific individuals, or represent a collective definition agreed upon
by many individuals.  As many dictionaries source their definitions
from other dictionaries, we have endeavoured to always list the
original source.

\begin{enumerate}

\item ``The ability to use memory, knowledge, experience,
  understanding, reasoning, imagination and judgement in order to
  solve problems and adapt to new situations.'' AllWords Dictionary,
  2006

\item ``The capacity to acquire and apply knowledge.'' The American
  Heritage Dictionary, fourth edition, 2000

\item ``Individuals differ from one another in their ability to
  understand complex ideas, to adapt effectively to the environment,
  to learn from experience, to engage in various forms of reasoning,
  to overcome obstacles by taking thought.''  American Psychological
  Association~\cite{Neisser:96}

\item ``The ability to learn, understand and make judgments or have
  opinions that are based on reason'' Cambridge Advance Learner's
  Dictionary, 2006

\item ``Intelligence is a very general mental capability that, among
  other things, involves the ability to reason, plan, solve problems,
  think abstractly, comprehend complex ideas, learn quickly and learn
  from experience.'' Common statement with 52 expert
  signatories~\cite{Gottfredson:97msoi}

\item ``The ability to learn facts and skills and apply them,
  especially when this ability is highly developed.'' Encarta World
  English Dictionary, 2006

\item ``\ldots ability to adapt effectively to the environment, either
  by making a change in oneself or by changing the environment or
  finding a new one \ldots intelligence is not a single mental
  process, but rather a combination of many mental processes directed
  toward effective adaptation to the environment.'' Encyclopedia
  Britannica, 2006

\item ``the general mental ability involved in calculating, reasoning,
  perceiving relationships and analogies, learning quickly, storing
  and retrieving information, using language fluently, classifying,
  generalizing, and adjusting to new situations.'' Columbia
  Encyclopedia, sixth edition, 2006

\item ``Capacity for learning, reasoning, understanding, and similar
  forms of mental activity; aptitude in grasping truths,
  relationships, facts, meanings, etc.''  Random House Unabridged
  Dictionary, 2006

\item ``The ability to learn, understand, and think about things.''
  Longman Dictionary or Contemporary English, 2006

\item ``: the ability to learn or understand or to deal with new or
  trying situations : \ldots\  the skilled use of reason (2) :
  the ability to apply knowledge to manipulate one's environment or to
  think abstractly as measured by objective criteria (as tests)''
  Merriam-Webster Online Dictionary, 2006

\item ``The ability to acquire and apply knowledge and skills.''
  Compact Oxford English Dictionary, 2006

\item ``\ldots the ability to adapt to the environment.'' World Book
  Encyclopedia, 2006

\item ``Intelligence is a property of mind that encompasses many
  related mental abilities, such as the capacities to reason, plan,
  solve problems, think abstractly, comprehend ideas and language, and
  learn.'' Wikipedia, 4 October, 2006

\item ``Capacity of mind, especially to understand principles, truths,
  facts or meanings, acquire knowledge, and apply it to practise; the
  ability to learn and comprehend.'' Wiktionary, 4 October, 2006

\item ``The ability to learn and understand or to deal with
  problems.''  Word Central Student Dictionary, 2006

\item ``The ability to comprehend; to understand and profit from
  experience.''  Wordnet 2.1, 2006

\item ``The capacity to learn, reason, and understand.''  Wordsmyth
  Dictionary, 2006

\end{enumerate}

\section{Psychologist definitions}

This section contains definitions from psychologists.  In some cases
we have not yet managed to locate the exact reference and would
appreciate any help in doing so.

\begin{enumerate}

\item ``Intelligence is not a single, unitary ability, but rather a
  composite of several functions. The term denotes that combination of
  abilities required for survival and advancement within a particular
  culture.'' A. Anastasi~\cite{Anastasi:92}

\item ``\ldots that facet of mind underlying our capacity to think, to
  solve novel problems, to reason and to have knowledge of the world."
  M. Anderson~\cite{Anderson:06}

\item ``It seems to us that in intelligence there is a fundamental
  faculty, the alteration or the lack of which, is of the utmost
  importance for practical life. This faculty is judgement, otherwise
  called good sense, practical sense, initiative, the faculty of
  adapting ones self to circumstances.''
  \mbox{A. Binet~\cite{Binet:1905}}

\item ``We shall use the term `intelligence' to mean the ability of
  an organism to solve new problems \ldots''
  W. V. Bingham~\cite{Bingham:37}

\item ``Intelligence is what is measured by intelligence tests.''
  E. Boring~\cite{Boring:23}

\item  ``\ldots a quality that is intellectual and not emotional or
  moral: in measuring it we try to rule out the effects of the child's
  zeal, interest, industry, and the like. Secondly, it denotes a
  general capacity, a capacity that enters into everything the child
  says or does or thinks; any want of 'intelligence' will therefore be
  revealed to some degree in almost all that he attempts;''
  C. L. Burt~\cite{Burt:57}

\item ``A person possesses intelligence insofar as he has learned,
  or can learn, to adjust himself to his environment.''  S. S. Colvin
  quoted in~\cite{Sternberg:00}

\item ``\ldots the ability to plan and structure one's behavior with
  an end in view.'' J. P. Das 

\item ``The capacity to learn or to profit by experience.''
  W. F. Dearborn quoted in~\cite{Sternberg:00}

\item ``\ldots in its lowest terms intelligence is present where the
  individual animal, or human being, is aware, however dimly, of the
  relevance of his behaviour to an objective.  Many definitions of
  what is indefinable have been attempted by psychologists, of which
  the least unsatisfactory are 1. the capacity to meet novel
  situations, or to learn to do so, by new adaptive responses and
  2. the ability to perform tests or tasks, involving the grasping of
  relationships, the degree of intelligence being proportional to the
  complexity, or the abstractness, or both, of the relationship.''
  J. Drever~\cite{Drever:52}

\item ``Intelligence A: the biological substrate of mental ability,
  the brains' neuroanatomy and physiology; Intelligence B: the
  manifestation of intelligence A, and everything that influences its
  expression in real life behavior; Intelligence C: the level of
  performance on psychometric tests of cognitive ability.''
  H. J. Eysenck.

\item ``Sensory capacity, capacity for perceptual recognition,
  quickness, range or flexibility or association, facility and
  imagination, span of attention, quickness or alertness in
  response.''  F. N. Freeman quoted in~\cite{Sternberg:00}

\item ``\ldots adjustment or adaptation of the individual to his
  total environment, or limited aspects thereof \ldots the capacity to
  reorganize one's behavior patterns so as to act more effectively and
  more appropriately in novel situations \ldots the ability to learn
  \ldots the extent to which a person is educable \ldots the ability
  to carry on abstract thinking \ldots the effective use of concepts
  and symbols in dealing with a problem to be solved \ldots''
  W. Freeman 

\item ``An intelligence is the ability to solve problems, or to
  create products, that are valued within one or more cultural
  settings.'' H. Gardner~\cite{Gardner:93}

\item ``\ldots performing an operation on a specific type of content
  to produce a particular product.'' J. P. Guilford 

\item ``Sensation, perception, association, memory, imagination,
  discrimination, judgement and reasoning.'' N. E. Haggerty quoted
  in~\cite{Sternberg:00}

\item ``The capacity for knowledge, and knowledge possessed.''
  V. A. C. Henmon \cite{Henmon:21}

\item ``\ldots cognitive ability.'' R. J. Herrnstein and
  C. Murray~\cite{Herrnstein:96}

\item ``\ldots the resultant of the process of acquiring, storing in
  memory, retrieving, combining, comparing, and using in new contexts
  information and conceptual skills.'' Humphreys 

\item ``Intelligence is the ability to learn, exercise judgment, and
  be imaginative.''  J. Huarte 

\item ``Intelligence is a general factor that runs through all types
  of performance.'' A.~Jensen

\item  ``Intelligence is assimilation to the extent that it
  incorporates all the given data of experience within its framework
  \ldots There can be no doubt either, that mental life is also
  accommodation to the environment. Assimilation can never be pure
  because by incorporating new elements into its earlier schemata the
  intelligence constantly modifies the latter in order to adjust them
  to new elements.'' J. Piaget~\cite{Piaget:63}

\item ``Ability to adapt oneself adequately to relatively new
  situations in life.''  R.~Pinter quoted in~\cite{Sternberg:00}

\item ``A biological mechanism by which the effects of a complexity of
  stimuli are brought together and given a somewhat unified effect in
  behavior.''  J. Peterson quoted in~\cite{Sternberg:00}

\item ``\ldots certain set of cognitive capacities that enable an
  individual to adapt and thrive in any given environment they find
  themselves in, and those cognitive capacities include things like
  memory and retrieval, and problem solving and so forth. There's a
  cluster of cognitive abilities that lead to successful adaptation to
  a wide range of environments.'' D. K. Simonton~\cite{Simonton:03}

\item ``Intelligence is part of the internal environment that shows
  through at the interface between person and external environment as
  a function of cognitive task demands.'' R. E. Snow quoted
  in~\cite{Slatter:01}

\item ``\ldots I prefer to refer to it as `successful intelligence.'
  And the reason is that the emphasis is on the use of your
  intelligence to achieve success in your life.  So I define it as
  your skill in achieving whatever it is you want to attain in your
  life within your sociocultural context --- meaning that people have
  different goals for themselves, and for some it's to get very good
  grades in school and to do well on tests, and for others it might be
  to become a very good basketball player or actress or musician.''
  R. J. Sternberg~\cite{Sternberg:03}


\item ``\ldots the ability to undertake activities that are
  characterized by (1) difficulty, (2) complexity, (3) abstractness,
  (4) economy, (5) adaptedness to goal, (6) social value, and (7) the
  emergence of originals, and to maintain such activities under
  conditions that demand a concentration of energy and a resistance to
  emotional forces.'' Stoddard 

\item ``The ability to carry on abstract thinking.''  L. M. Terman
  quoted in~\cite{Sternberg:00}

\item ``Intelligence, considered as a mental trait, is the capacity to
  make impulses focal at their early, unfinished stage of formation.
  Intelligence is therefore the capacity for abstraction, which is an
  inhibitory process.''  L. L. Thurstone~\cite{Thurstone:24}

\item ``The capacity to inhibit an instinctive adjustment, the
  capacity to redefine the inhibited instinctive adjustment in the
  light of imaginally experienced trial and error, and the capacity to
  realise the modified instinctive adjustment in overt behavior to
  the advantage of the individual as a social animal.''
  L. L. Thurstone quoted in~\cite{Sternberg:00}

\item ``A global concept that involves an individual's ability to
  act purposefully, think rationally, and deal effectively with the
  environment.''  D. Wechsler~\cite{Wechsler:58}

\item ``The capacity to acquire capacity.''  H. Woodrow quoted
  in~\cite{Sternberg:00}

\item ``\ldots the term intelligence designates a complexly
  interrelated assemblage of functions, no one of which is completely
  or accurately known in man \ldots'' R. M. Yerkes and
  A. W. Yerkes~\cite{Yerkes:29}

\item ``\ldots that faculty of mind by which order is perceived in a
  situation previously considered disordered.''  R. W. Young quoted
  in~\cite{Kurzweil:00}

\end{enumerate}

\section{AI researcher definitions}

This section lists definitions from researchers in artificial
intelligence.

\begin{enumerate}

\item ``\ldots the ability of a system to act appropriately in an
  uncertain environment, where appropriate action is that which
  increases the probability of success, and success is the achievement
  of behavioral subgoals that support the system's ultimate goal.''
  J. S. Albus~\cite{Albus:91}

\item ``Any system \ldots that generates adaptive behviour to meet
  goals in a range of environments can be said to be intelligent.''
  D. Fogel~\cite{Fogel:95}

\item ``Achieving complex goals in complex environments''
  B. Goertzel~\cite{Goertzel:06}

\item ``Intelligent systems are expected to work, and work well, in
  many different environments. Their property of intelligence allows
  them to maximize the probability of success even if full knowledge
  of the situation is not available.  Functioning of intelligent
  systems cannot be considered separately from the environment and the
  concrete situation including the goal.''
  R. R. Gudwin~\cite{Gudwin:00}

\item ``[Performance intelligence is] the successful (i.e.,
  goal-achieving) performance of the system in a complicated
  environment.'' J. A. Horst~\cite{Horst:02}

\item ``Intelligence is the ability to use optimally limited
  resources -- including time -- to achieve goals.''
  R. Kurzweil~\cite{Kurzweil:00}

\item ``Intelligence is the power to rapidly find an adequate solution
  in what appears \emph{a priori} (to observers) to be an immense
  search space.'' D. Lenat and E. Feigenbaum~\cite{Lenat:91}

\item ``Intelligence measures an agent's ability to achieve goals in a
  wide range of environments.'' S. Legg and
  M. Hutter~\cite{Legg:06ior}

\item ``\ldots doing well at a broad range of tasks is an empirical
  definition of `intelligence'$\;$'' H. Masum~\cite{Masum:02}

\item ``Intelligence is the computational part of the ability to
  achieve goals in the world. Varying kinds and degrees of
  intelligence occur in people, many animals and some machines.''
  J. McCarthy~\cite{McCarthy:04}

\item ``\ldots the ability to solve hard problems.''
  M. Minsky~\cite{Minsky:85}

\item ``Intelligence is the ability to process information properly in
  a complex environment.  The criteria of properness are not
  predefined and hence not available beforehand. They are acquired as
  a result of the information processing.'' H.
  Nakashima~\cite{Nakashima:99}

\item ``\ldots in any real situation behavior appropriate to the
  ends of the system and adaptive to the demands of the environment
  can occur, within some limits of speed and complexity.'' A. Newell
  and H. A. Simon~\cite{Newell:76}

\item ``[An intelligent agent does what] is appropriate for its
  circumstances and its goal, it is flexible to changing environments
  and changing goals, it learns from experience, and it makes
  appropriate choices given perceptual limitations and finite
  computation.'' D. Poole~\cite{Poole:98}

\item ``Intelligence means getting better over time.''
  Schank~\cite{Schank:91}

\item ``Intelligence is the ability for an information processing
  system to adapt to its environment with insufficient knowledge and
  resources.'' P. Wang~\cite{Wang:95}

\item ``\ldots the mental ability to sustain successful life.''
  K. Warwick quoted in~\cite{Asohan:03}

\item ``\ldots the essential, domain-independent skills necessary
  for acquiring a wide range of domain-specific knowledge -- the
  ability to learn anything.  Achieving this with `artificial general
  intelligence' (AGI) requires a highly adaptive, general-purpose
  system that can autonomously acquire an extremely wide range of
  specific knowledge and skills and can improve its own cognitive
  ability through self-directed learning.'' P. Voss~\cite{Voss:05}

\end{enumerate}

\section{Is a single definition possible?}

In matters of definition, it is difficult to argue that there is an
objective sense in which one definition could be considered to be the
correct one.  Nevertheless, some definitions are clearly more concise,
precise and general than others.  Furthermore, it is clear that many
of the definitions listed above are strongly related to each other and
share many common features.  If we scan through the definitions
pulling out commonly occurring features we find that intelligence:

\begin{itemize}

\item Is a property that an individual agent has as it interacts with its
  environment or environments.

\item Is related to the agent's ability to succeed or profit with
  respect to some goal or objective.

\item Depends on how able the agent is to adapt to different objectives
  and environments.

\end{itemize}

Putting these key attributes together produces the informal definition
of intelligence that we have adopted,

\begin{quote}
``Intelligence measures an agent's ability to achieve goals in a wide
  range of environments.'' S. Legg and M. Hutter
\end{quote}

Features such as the ability to learn and adapt, or to understand, are
implicit in the above definition as these capacities enable an agent
to succeed in a wide range of environments.  For a more comprehensive
explanation, along with a mathematical formalisation of the above
definition, see~\cite{Legg:06ior} or our forthcoming journal paper.


\end{document}